%% file: main.tex
\documentclass[final]{cvpr}
\pdfoutput=1
\usepackage{times}
\usepackage{epsfig}
\usepackage{graphicx}
\usepackage{amsmath}
\usepackage{amssymb}
\usepackage{multicol}
\usepackage{enumitem}
\usepackage{verbatim}
\usepackage{subfiles}
\usepackage[pagebackref=true,breaklinks=true,colorlinks,bookmarks=false]{hyperref}
\begin{document}

\title{\ SIPSA-Net: Shift-Invariant Pan Sharpening with Moving Object Alignment for Satellite Imagery}

\author{Jaehyup Lee \qquad\qquad Soomin Seo \qquad\qquad Munchurl Kim\thanks{Corresponding author.}\\ \\
[0.5em]
{Korea Advanced Institue of Science and Technology (KAIST)} \\
{\tt\small \{woguq365, ssm9462, mkimee\} @kaist.ac.kr}
}
\maketitle

\begin{abstract}
Pan-sharpening is a process of merging a high-resolution (HR) panchromatic (PAN) image and its corresponding low-resolution (LR) multi-spectral (MS) image to create an HR-MS and pan-sharpened image. However, due to the different sensors' locations, characteristics and acquisition time, PAN and MS image pairs often tend to have various amounts of misalignment. Conventional deep-learning-based methods that were trained with such misaligned PAN-MS image pairs suffer from diverse artifacts such as double-edge and blur artifacts in the resultant PAN-sharpened images. In this paper, we propose a novel framework called shift-invariant pan-sharpening with moving object alignment (SIPSA-Net) which is the first method to take into account such large misalignment of moving object regions for PAN sharpening. The SISPA-Net has a feature alignment module (FAM) that can adjust one feature to be aligned to another feature, even between the two different PAN and MS domains. For better alignment in pan-sharpened images, a shift-invariant spectral loss is newly designed, which ignores the inherent misalignment in the original MS input, thereby having the same effect as optimizing the spectral loss with a well-aligned MS image. Extensive experimental results show that our SIPSA-Net can generate pan-sharpened images with remarkable improvements in terms of visual quality and alignment, compared to the state-of-the-art methods. 
\end{abstract}

\section{Introduction}
The satellite images are being used in a wide range of applications such as environmental monitoring, surveillance systems, and mapping services as well. Google Earth$^\text{TM}$ is one of the most popular virtual globe applications. Such high-resolution multi-spectral images are obtained from commercialized pan-sharpening software that fuses low-resolution (LR) multi-spectral (MS) images and high-resolution (HR) single channel panchromatic (PAN) images to generate pan-sharpened (PS) images. The produced PS images should have similar high-frequency details as the PAN images and similar colors as the MS images. However, when the PS images in the virtual globe applications are compared with the original PAN and MS images, many types of artifacts are often observed for which the details of PS images are not as sharp as those of the PAN images and colors appear to be distorted from the MS images.

\begin{figure}[!t]
\centering
\includegraphics[width=1\columnwidth]{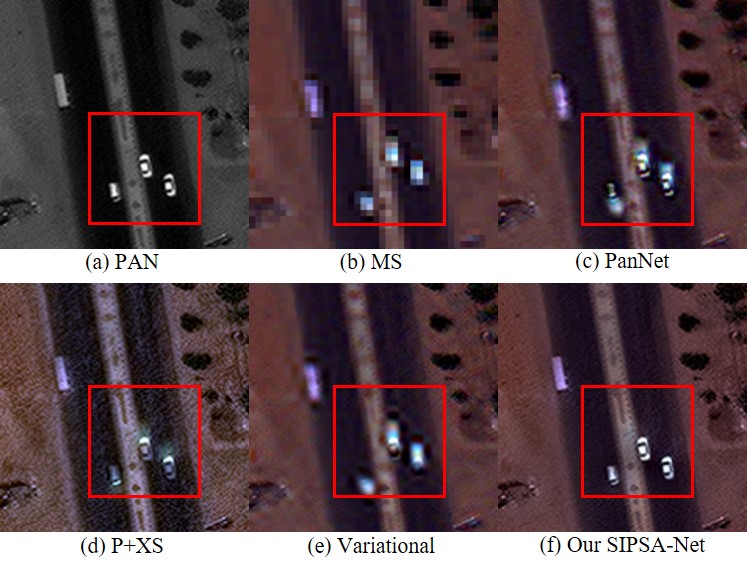}
\caption{Pan-sharpening results from different methods on WorldView-3 dataset. Our proposed SIPSA-Net generates an artifact-free pan-sharpened image, where other methods generate blurry and distorted images.}
\label{fig:first_figure}
\end{figure}

Recently, many deep-learning-based pan-sharpening methods \cite{s3, dsen2,pnn,pannet,bdpn} have been proposed and shown to outperform previous state-of-the-art pan-sharpening methods \cite{Regularize,pxs,variational,decomposition,undecimated}. Most of these methods train their networks by minimizing the difference between the pseudo-ground-truth MS images and the network output PS images in supervised manners, despite of the absence of the actual ground-truth PS images. One of the main difficulties of the pan-sharpening is the inherent misalignment between PAN and MS image pairs. Due to the different sensors' characteristics, physical locations and capturing time, PAN and MS image pairs inevitably have large pixel-misalignment that is even worse for locally moving objects such as cars on a highway. This often leads to various artifacts in output PS images such as double edges and blur artifact especially on the moving object regions having extreme misalignment.

In this paper, we propose a novel framework called shift-invariant pan-sharpening (SIPSA-Net) that solves the PS problem considering the global misalignment as well as the local misalignment due to different acquisition time for moving objects on the ground. SIPSA-Net is optimized to align the colors from MS images to the corresponding shapes in PAN images through a newly proposed feature alignment module (FAM). SIPSA-Net further elaborates the alignment by utilizing our novel shift-invariant spectral loss that ignores the inherent misalignment in the original MS input, thereby having the same effect as optimizing the spectral loss with an well-aligned MS image. As shown in Fig. \ref{fig:first_figure}, the output image from SIPSA-Net shows good alignment between structures and colors especially around the moving cars. In the PS outputs from other methods, the colors of the cars are smeared in the upward direction as comet tail artifacts. All source codes are publicly available at \url{https://github.com/brachiohyup/SIPSA}.

\subsection{Our contributions}

Our contributions can be summarized as follows:

\begin{itemize}[leftmargin=5pt,itemsep=0.3pt,topsep=0.3pt]
\item[$\bullet$] Alignment-aware pan-sharpening: Except for the global registration, none of the previous PS methods considered the extreme local misalignment induced by moving objects. We propose the first deep-learning-based PS method that can generate both locally and globally well-aligned PS images from misaligned PAN-MS image pairs.
\item[$\bullet$] Feature alignment module: The newly proposed feature alignment module learns a probability map of offsets in the feature domain for the aligned MS pixels with respect to the pixel locations in the PAN image to cope with both global and local misalignment. The probability map and the features extracted from the MS image are then used to generate the aligned MS image with respect to its corresponding PAN image. 
\item[$\bullet$] Shift-invariant spectral (SiS) loss: The conventional deep-learning-based PS methods have optimized their networks by using the misaligned MS images as the pseudo-ground-truth. However, this approach leads to artifacts in the PS outputs such as double-edge and blur artifacts. To remedy this, a SiS loss is proposed, which is effective in transferring the spectral information from the misaligned MS images to match the corresponding details in PAN images. SiS loss is calculated as the minimum difference between the output PS image and each of the multiple-shifted versions of MS input images. In this way, the loss becomes shift-invariant in color registration of the misaligned MS image to the PAN image.
\end{itemize}

\section{Related work}
\subsection{Traditional pan-sharpening methods}

Many conventional PS methods have been developed over the last decades, such as multi-resolution analysis methods \cite{decomposition,undecimated}, component substitution algorithms \cite{ihs,brovey,pca}, and model-based algorithms \cite{Regularize,pxs,variational}.

The multi-resolution analysis (MRA) methods \cite{decomposition,undecimated} are based on the idea that the PAN images have high-frequency details which are not present in MS images. Such high-frequency details are separated from the PAN images using multi-scale decomposition techniques and then adaptively fused with the upsampled version of the MS images. 

The component substitution (CS) algorithms separate the spatial and spectral information of the MS images using a specific transformation, and then the spatial component is substituted with the PAN image.  The CS-based algorithms include pan-sharpening methods based on an intensity hue-saturation technique \cite{ihs}, principal component analysis \cite{pca}, and Brovey transforms \cite{brovey}. These methods are extremely fast because they only need to upscale the MS input image and apply some spectral transformation to separate and replace the spatial component. 
\begin{figure*}[tb]
  \centering 
\includegraphics[width=1\textwidth]{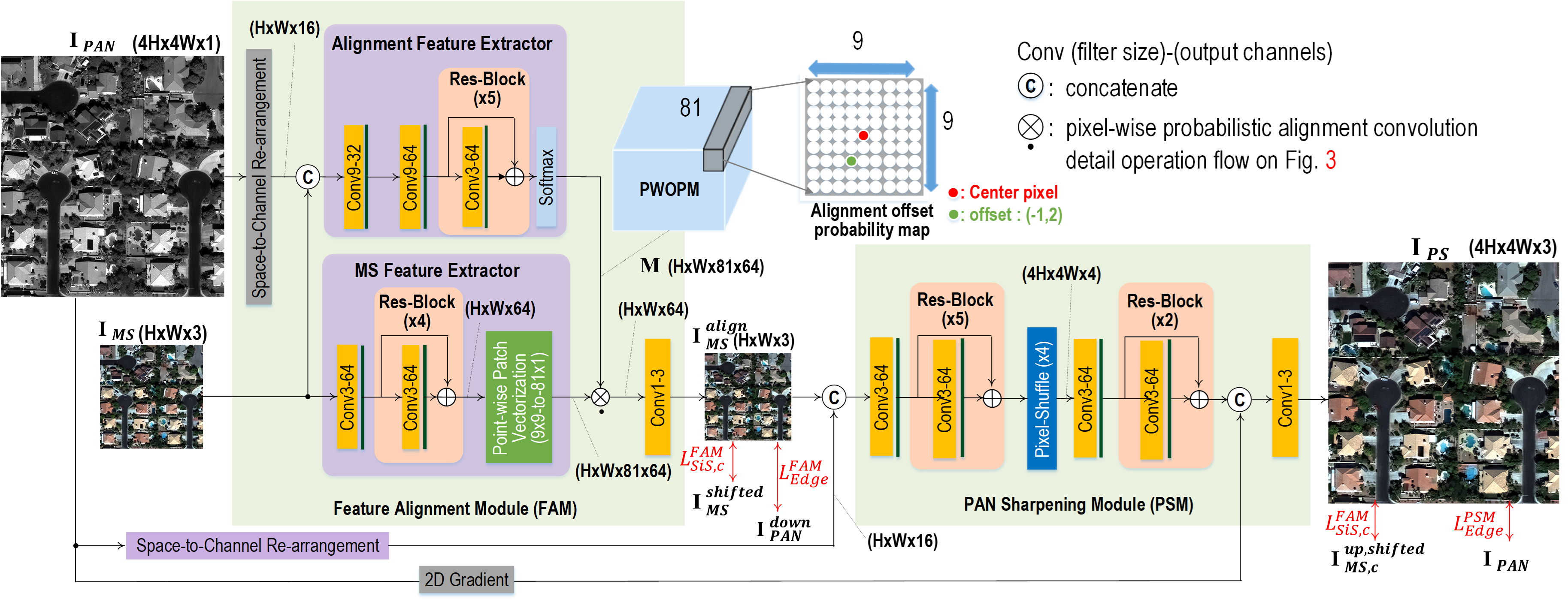}
  \caption{Architecture of our two-stage shift-invariant pan-sharpening (SIPSA-Net): alignment and pan-sharpening. The SIPSA-Net is trained with two losses: an edge detail loss for spatial detail preservation and a shift-invariant loss for spectral information preservation.}
  \label{fig:architecture}
\end{figure*}

The model-based algorithms \cite{Regularize,pxs,variational} treat the PAN and MS images as the spectral and spatial degraded versions of the ideal PS images. Under such assumption, pan-sharpening becomes a restoration problem aiming to reconstruct the ideal PS images from their degraded observations (PAN and MS input). Model-based algorithms optimize an objective function designed with some prior knowledge that is independent of specific training data. The methods require high computational complexity compared to the previously mentioned methods due to the optimization process.

\subsection{Deep-learning based methods}
Recently, many deep-learning-based pan-sharpening methods  \cite{s3,dsen2,pnn,pannet,bdpn} have been proposed. The architectures of such methods are based on convolutional neural networks (CNNs). Many types of CNN architectures for pan-sharpening have been proposed, and the proposed methods have shown a large margin of performance improvements over the conventional methods.

The goal of the pan-sharpening task is to obtain high-resolution PS images. Without the PAN images as guidance, this solely becomes a super-resolution (SR) task. Therefore, most previous deep-learning-based pan-sharpening methods have borrowed the CNN architectures of existing SR frameworks \cite{srcnn,edsr,rcan}. However, to make good use of the high-resolution PAN images, loss functions should be carefully designed to help pan-sharpening networks maintain the high-frequency details from PAN images and spectral information from MS images.

The first deep-learning-based pan-sharpening method is PNN \cite{pnn}. The network of PNN is based on the SRCNN \cite{srcnn}, the first CNN-based SR method. The network of PNN is composed of three convolutional layers. It is optimized to minimize the difference between lower-scaled pan-sharpened images and target MS images of the original-scale, which are given pseudo-ground-truths. Yang \textit{et al.} proposed PanNet \cite{pannet} which trains their network in the high-pass filtered domain rather than the image domain for better spatial structure preservation. DSen2 \cite{dsen2} shares a similar idea with PanNet in their network design and loss functions. The network architectures of both PanNet and DSen2 are based on an SR network called VDSR \cite{vdsr}, which have improved the SR quality over SRCNN. Zhang \textit{et al.} proposed BDPN \cite{bdpn} which utilizes a bi-directional network design to fuse PAN and MS images.

However, there are some difficulties when generating PS images from misaligned PAN-MS image pairs. S3 \cite{s3} is the first paper that considers the artifacts that come from such misalignment. S3 proposes spectral and spatial loss functions based on the correlation maps between MS and PAN inputs. The correlation maps are used as weights for the two-loss functions. They put more weights on the spectral loss function for the areas with higher correlation values, and those having lower correlation values have higher weights for the spatial loss function. This approach has improved the visual quality of PS outputs by reducing the artifacts from the misalignment. However, S3 has a critical drawback that the areas with lower correlation values (regions with large misalignment) are mostly affected by the spatial loss. Therefore the original colors from MS images are diminished. An appropriate way to handle the misalignment would be moving the color of an object on the MS image to match its shape on the corresponding PAN image.

\begin{figure*}[tb]
  \centering 
\includegraphics[width=1\textwidth]{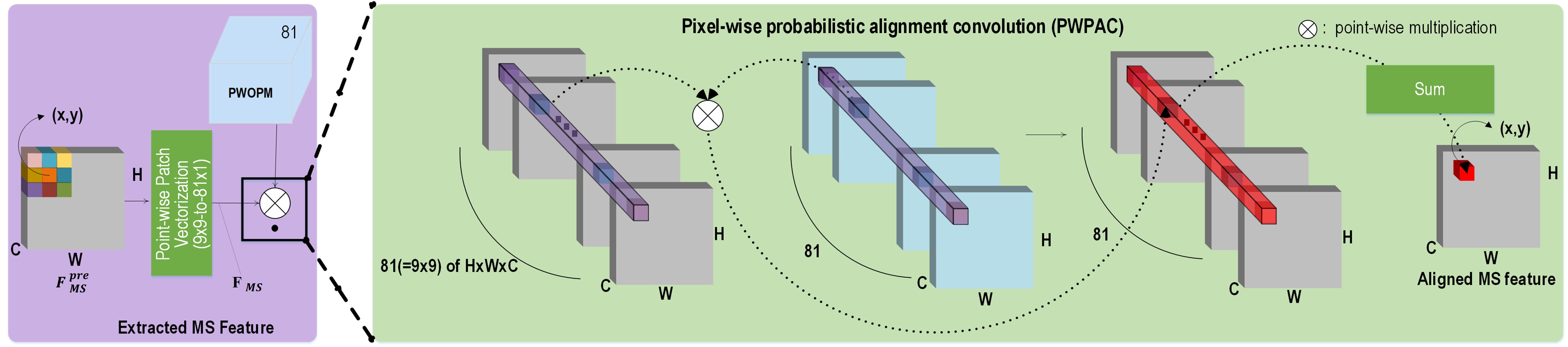}
  \caption{An illustration of pixel-wise probabilistic alignment convolution (PWPAC). The PWPAC operates depth (or channel)-wise multiplications of the PWOPM $\textbf{M}^P$ and the rearranged MS feature maps $\textbf{F}_{MS}$, and sums them up along each channel, yielding an aligned feature value $F_{MS}^{align}$$(x,y)$.}
  \label{fig:fam}
\end{figure*}

\section{Method}
\label{sec:method}
Fig. \ref{fig:architecture} illustrates the architecture of our proposed shift-invariant pan-sharpening network (SIPSA-Net). SIPSA-Net has a two-stage pan-sharpening structure with a feature alignment module (FAM) and a pan-sharpening module (PSM). FAM corrects the misaligned MS images to be aligned such that the colors of the MS images are matched with the corresponding shapes in the PAN images. PSM generates PS images from the aligned MS images and the PAN images. The modules are trained in an end-to-end manner with the shift-invariant loss function and edge loss function applied for the aligned MS output and PS output.

\subsection{Feature alignment module}
\label{sec:alignment stage}
As mentioned earlier, one of the main limitations of the previous CNN-based pan-sharpening methods is that their networks are simply trained with L1 or L2 loss functions for the misaligned MS-PAN image pairs without any registration \textit{a priori}. This results in the PS output with various artifacts such as color bleed and double edges. However, direct alignment between the PAN and MS images are very difficult due to their domain discrepancy with different signal characteristics and resolutions. Instead, the alignment might be better solved in the feature domain. So, we propose a novel feature alignment module (FAM) that aligns each MS input image to its corresponding PAN image in the feature domain, yielding an aligned MS image that is later fed into the pan-sharpening module (PSM) of the SIPSA-Net. As shown in Fig. \ref{fig:architecture}, the resulting aligned MS image $\textbf{I}^{align}_{MS}$ has the same size as the input MS image $\textbf{I}_{MS}$ of size \textit{W}$\times$\textit{H}$\times$3 and is aligned to its corresponding downscaled PAN image $\textbf{I}^{down}_{PAN}$ of size \textit{W}$\times$\textit{H}. The alignment in the FAM helps the PSM easily generate the robust and stable PS output $\textbf{I}_{PS}$ with a good alignment between colors and shapes, given an input MS-PAN pair with large misalignment.

As shown in Fig. \ref{fig:architecture}, the FAM includes two feature extractors: alignment and MS feature extractors. The alignment feature extractor learns a pixel-wise offset probability map (PWOPM) of size W$\times$H$\times$81 to align the features $\textbf{F}_{MS}$ of MS input $\textbf{I}_{MS}$ to the features $\textbf{F}_{PAN}$ of the corresponding PAN input $\textbf{I}_{PAN}$ of size 4\textit{W}$\times$4\textit{H}$\times$1, thus yielding the aligned MS features $\textbf{F}^{align}_{MS}$. The offset value represents the amount of misalignment between PAN and MS features, which can be further utilized to align two features in following operations. The MS feature extractor converts $\textbf{I}_{MS}$ in pixel-domain to $\textbf{F}_{MS}$ in feature-domain. The output of the FAM for the concatenated input of $\textbf{I}_{MS}$ and the space-to-channel-rearranged PAN input $\textbf{I}^{s2c}_{PAN}$ of size \textit{W}$\times$\textit{H}$\times$16 is the PWOPM where a 9$\times$9-sized alignment offset probability map is positioned at each pixel location along the 81-depth channel dimension. Each 9$\times$9-sized alignment offset probability map is obtained via a softmax function, which is then convolved with the 9$\times$9-sized local region centered at the corresponding pixel location. Note that all the feature channels share the 9$\times$9-sized alignment offset probability map at the same pixel locations. By doing so, the aligned MS features $\textbf{F}^{align}_{MS}$ are obtained.

As illustrated in the right upper part of Fig. \ref{fig:architecture}, an 81-dimensional feature vector in the PWOPM is shown in a rearranged 9$\times$9-sized offset probability map where the red circle indicate the center pixel location (0,0) and the green circle in the (-1, 2) location has the highest offset probability value. This indicates that the MS feature at the (-1, 2) location is likely to be moved to the center pixel location. The aforementioned FAM operation is expressed in Eq. \ref{PWPAC}.
Let the PWOPM be $\textbf{M} = [M_{0,0}, ...,M_{W,H}]$, $M_{x,y} \in \mathcal{R}^{9 \times 9}$. By the pixel-wise probabilistic alignment convolution (PWPAC), the extracted MS feature $\textbf{F}_{MS}$, $\textbf{F}_{MS} \in \mathcal{R}^{W \times H \times C}$ can be point-wise aligned to $\textbf{I}^{down}_{PAN}$. The resulting aligned \textit{c}-th feature map can be expressed as 
\begin{equation}
\label{PWPAC}
    \begin{aligned}
\textbf{F}^{align,(x,y)}_{MS,c} = \sum_{i=-4}^{4} \sum_{j=-4}^{4} \textbf{F}_{MS,c}^{(x+i,y+j)} \cdot M^{(i,j)}_{x,y}
\end{aligned}
\end{equation}
where $1\leq x\leq W, 1\leq y\leq H$, and $1\leq c\leq C$. It should be noted that our PWPAC is distinguished from pixel-wise adaptive convolution (PAC) \cite{pac}. The PAC operates local convolution with learned filter values for up-sampling while the PWPAC performs pixel-wise alignment in a probabilistic sense, \textit{i.e.}, probabilistic alignment convolution, for the extracted features of an MS input with respect to the local features of the PAN input.

Fig. \ref{fig:fam} shows the feature extraction for the MS input by the MS feature extractor. The MS feature extractor extracts a feature map $F^{\textit{pre}}_{MS}$ and expands it to a set of feature maps, $\textbf{F}_{MS}$, via point-wise patch vectorization operations that rearrange each 9$\times$9-sized patch in stride 1 into the channel dimension. $\textbf{F}_{MS}$ is a set of point-wise shifted versions of $F^{\textit{pre}}_{MS}$ from (-4,-4) to (4,4). Therefore, the shifted version by (0, 0), which is the middle channel, is identical to $F^{\textit{pre}}_{MS}$. In this way, Eq. \ref{PWPAC} performs probabilistic alignment operation over the set of shifted versions of $F^{\textit{pre}}_{MS}$ by PWOPM $\textbf{M}$.    

\subsection{Pan-sharpening module}
\label{sec:pan-sharpening stage}
The pan-sharpening module (PSM) in Fig. \ref{fig:architecture} performs the pan-sharpening based on the aligned MS input $\textbf{I}^{align}_{MS}$ and the space-to-channel-rearranged PAN input $\textbf{I}^{s2c}_{PAN}$ of size \textit{W}$\times$\textit{H}$\times$16. After the first five ResBlocks, the intermediate output is up-scaled 4 times via the Pixel-Shuffle layer. The 2D gradient of $\textbf{I}_{PAN}$ is concatenated to the output of the second two ResBlocks, which are then fed as input into the last convolution layer. It is worthwhile to note that the 2D gradient helps maintain the details of the PAN input $\textbf{I}_{PAN}$ while producing the final PS output $\textbf{I}_{PS}$.

\subsection{Loss function}
\label{sec:loss}
For the training of SIPSA-Net, two different types of loss functions are utilized to generate accurate pan-sharpened images. For this, the edge detail loss and the shift-invariant spectral loss are incorporated for spatial and spectral information preservation, respectively. 

\paragraph{Edge detail loss}
\label{sec:edge detail loss}
The edge detail loss is designed to preserve the edge details of the PAN images when generating the PS images. We set the ground truth of the spatial details of the output PS image as the edge map of input PAN image. It should be noted that, due to the different characteristics of the PAN and MS image signals, the edge directions may be different, although their positions are the same. Therefore, we utilize the absolute edge values of the PAN images as ground truth for those of the luminance of the PS output. The two edge losses are computed for FAM and PSM modules, which are given by :
\begin{equation}
\label{align edge}
    \begin{aligned}
        L^{FAM}_{Edge} = \parallel{|{\nabla{I^{\textit{lum,align}}_{MS}}|-|\nabla I_{PAN}^{down}}}|\parallel
    \end{aligned}
\end{equation}
\begin{equation}
\label{ps edge}
    \begin{aligned}
        L_{Edge}^{PSM} = {\parallel{|\nabla I^{\textit{lum}}_{PS}|-|\nabla I_{PAN}}}|\parallel
    \end{aligned}
\end{equation}
where $I^{\textit{lum,align}}_{MS}$ and $I^{\textit{lum}}_{PS}$ are the luminance components of $I^{align}_{MS}$ and $I_{PS}$ computed as the average of the RGB channels, respectively, and $\nabla$ is a 2-D gradient operator. 

\begin{table*}[tb]
    \small
    \centering
    \setlength{\tabcolsep}{5pt}
    \caption{Quantitative comparison (measured with original misaligned MS and aligned MS images).}
    \input{with_psnr}
    \label{tab:metricQualitative comparisons_comparison}
    
\end{table*}

\paragraph{Shift-invariant spectral loss (SiS loss)}
\label{sec:shift-invariant loss}
The conventional spectral loss minimizes \cite{dsen2,pnn,pannet,bdpn} the difference between the down-scaled PS images and the input MS images without alignment, thus leading the PS networks to produce comet tail artifacts for color, especially for objects with large misalignment.

To resolve the issue, we propose two shift-invariant spectral (SiS) losses: $(i)$ one is aimed at minimizing the minimum among the differences between the aligned MS image and each of multiple-shifted versions of an MS input image for the FAM, denoted as $L^{FAM}_{SiS}$ ; and $(ii)$ the other one is used to minimize the minimum among the differences between the PS output image and each of multiple-shifted versions of an up-scaled MS input image for the PSM, denoted as $L^{PSM}_{SiS}$. The two SiS losses have the same effects as optimizing the spectral losses with well-aligned MS input images. It is possible to precisely transfer color information from the MS input images to the output PS images, regardless of misalignment between MS and PAN images. Our $c$-channel SiS losses for $L^{FAM}_{SiS}$ and $L^{PSM}_{SiS}$ are given by : 
\begin{equation}
\label{eq:SIS_ams}
    \begin{aligned}
        L_{SiS,c}^{FAM}(x,y)  =  \min_{-4\le i,j\le4} | I_{MS,c}^{align,(x,y)} - I_{MS,c}^{(x+i,y+j)}|
    \end{aligned}
\end{equation}
\begin{equation}
\label{eq:SIS_ps}
    \begin{aligned}
        L_{SiS,c}^{PSM}(x,y)= \min_{-4\le i,j \le4} |I_{PS,c}^{(x,y)} - I_{MS,c}^{up,(x+4i,y+4j)}|
    \end{aligned}
\end{equation}
where $i$ and $j$ are integer values, and $\textbf{I}_{MS}^{up}$ is the upscaled input MS image by bilinear interpolation. $I_{MS,c}^{align,(x,y)}$, $I_{MS,c}^{(x,y)}$ and $I_{MS,c}^{up,(x,y)}$ are the $c$-th channel components of $\textbf{I}_{MS}^{align}$, $\textbf{I}_{MS}$ and $\textbf{I}_{MS}^{up}$ at location $(x,y)$, respectively. In Eqs. \ref{eq:SIS_ams} and \ref{eq:SIS_ps}, the shift range is set to 9$\times$9 with stride 1 for $\textbf{I}_{MS}^{align}$ and 36$\times$36 with stride 4 for $\textbf{I}_{PS}$, because the amount of misalignment between the PAN and MS images are less than 3 pixels in MS scale for most of the cases in WorldView-3 dataset. The shift range should be adjusted for different dataset with larger misalignment with prior knowledge, which is not problematic because different satellite image sets have different signal characteristics and require their own dedicated networks for training and testing as frequently done in the previous methods. The SiS loss for the PSM is effective in learning the color from the misaligned ground truths of upscaled MS images $\textbf{I}_{MS}^{up}$ and transferring them to the PS images $\textbf{I}_{PS}$ in a shift-invariant manner. 

The total loss function for training the proposed SIPSA-Net is defined as a weighted sum of the aforementioned loss functions, which is given by : 
\begin{equation}
\label{eq:loss_total}
    \begin{aligned}
        L_{total}  = 4L^{FAM}_{SiS}+ L^{PSM}_{SiS}+\alpha  ({4L^{FAM}_{Edge}}+L^{PSM}_{Edge})
    \end{aligned}
\end{equation}
where $\alpha$ is empirically set to 2. The SiS losses are computed as
$L^{FAM}_{SiS} = \sum_{x=1}^{W} \sum_{y=1}^{H} \sum_{c=1}^{3} L_{SiS,c}^{FAM}(x,y) $ and $L^{PSM}_{SiS} = \sum_{x=1}^{4W} \sum_{y=1}^{4H} \sum_{c=1}^{3} L_{SiS,c}^{PSM}(x,y) $.

\section{Experiment}
\subsection{Experiment Settings}
\paragraph{Datasets}
All the previous methods including ours and baselines were trained and tested on the WorldView-3 satellite image dataset provided by DigitalGlobe$^\text{TM}$. Multi-spectral (MS) images are composed of 8 bands: red, red edge, coastal, blue, green, yellow, near-IR1 and near-IR2 in the unit of 1.24$m$ ground sample distance (GSD). Their corresponding panchromatic (PAN) images are single-channel images of 0.31$m$ GSD. The dynamic range is 11-bit per pixel for both PAN and MS images. We used 11,799 MS-PAN image pairs for training SIPSA-Net. For training, we randomly cropped the MS and PAN images to the patch pairs of 128$\times$128 and 512$\times$512 sizes, respectively. The input MS and PAN images are then normalized to have a value between 0 and 1 before being fed into the networks.

\paragraph{Training details}
We trained all the networks by the decoupled ADAMW \cite{adamw} optimizer with an initial learning rate of $10^{-4}$ and initial weight decay of $10^{-7}$. The networks were trained for total $10^6$ iterations, where the learning rate and weight decay were lowered by a factor of 10 after 5$ \times10^5$ iterations. The mini-batch size was set to 2. All the deep-learning based pan-sharpening methods including our methods were implemented with Tensorflow \cite{tensorflow} and were trained and tested on NVIDIA Titan$^\text{TM}$ Xp GPU. The non-deep learning methods were tested on a server platform with Intel Xeon CPU E5-2637 and 128GB RAM.

\subsection{Comparisons and Discussions}
The results of the proposed SIPSA-Net are compared with (i) \textit{seven non-deep-learning PS methods} including Brovey transform \cite{brovey}, affinity PS \cite{affinity}, guided-filtering-based PS \cite{guided}, intensity-hue-saturation (IHS) PS \cite{ihs}, principal component analysis (PCA) PS \cite{pca}, P+XS PS \cite{pxs} and variational PS \cite{variational}, and (ii) \textit{five deep-learning-based PS methods} including PNN \cite{pnn}, PanNet \cite{pannet}, DSen2 \cite{dsen2}, and their variants trained with S3 loss \cite{s3}, called PanNet-S3 and DSen2-S3, respectively. For testing, we randomly selected 100 PAN-MS image pairs in the WorldView-3 test dataset.
\begin{figure*}[!]
  \centering 
\includegraphics[width=0.9\textwidth]{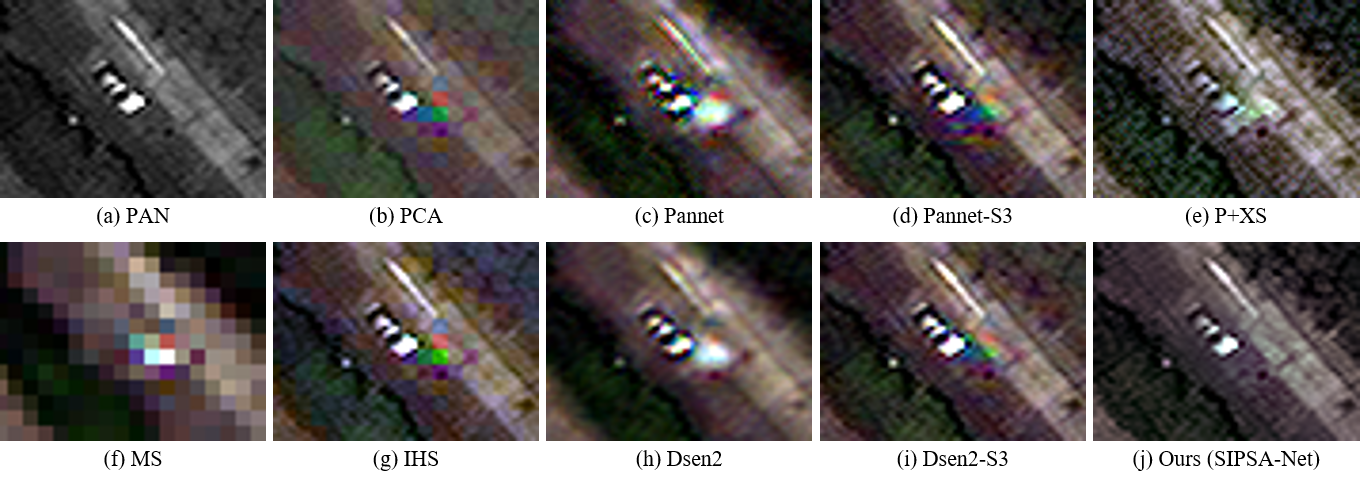}
  \caption{Result images for pan-sharpening using various method and out SIPSA-Net.}
  \label{fig:fig_4}
\end{figure*}
\begin{figure*}[!]
  \centering 
\includegraphics[width=0.9\textwidth]{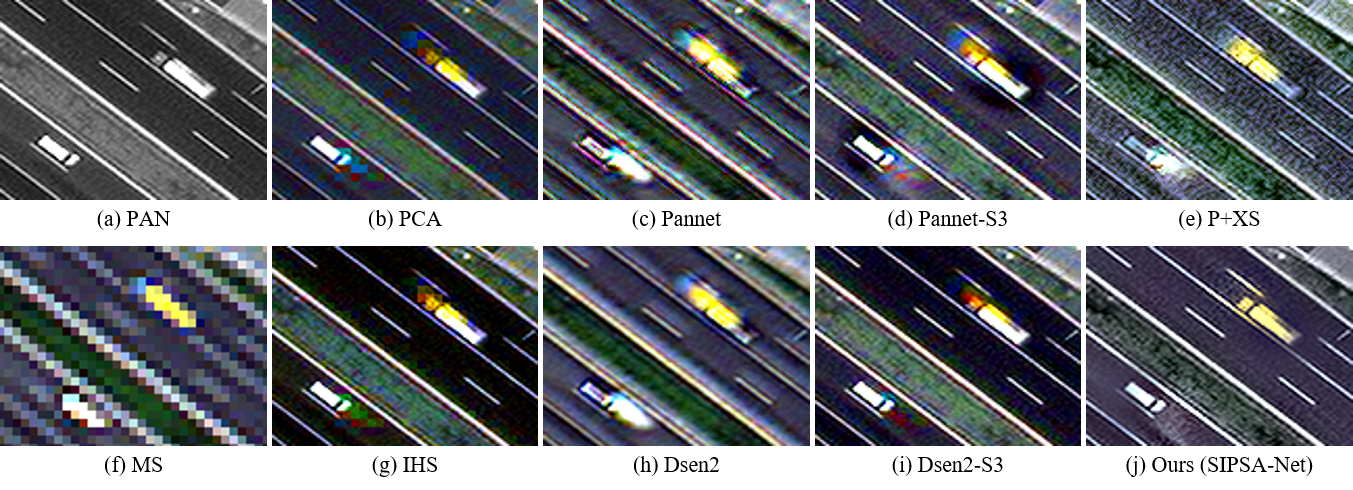}
  \caption{Result images for pan-sharpening using various method and out SIPSA-Net from extremely misaligned MS-PAN image.}
  \label{fig:fig_5}
\end{figure*}
\begin{figure*}[!]
  \centering 
\includegraphics[width=0.9\textwidth]{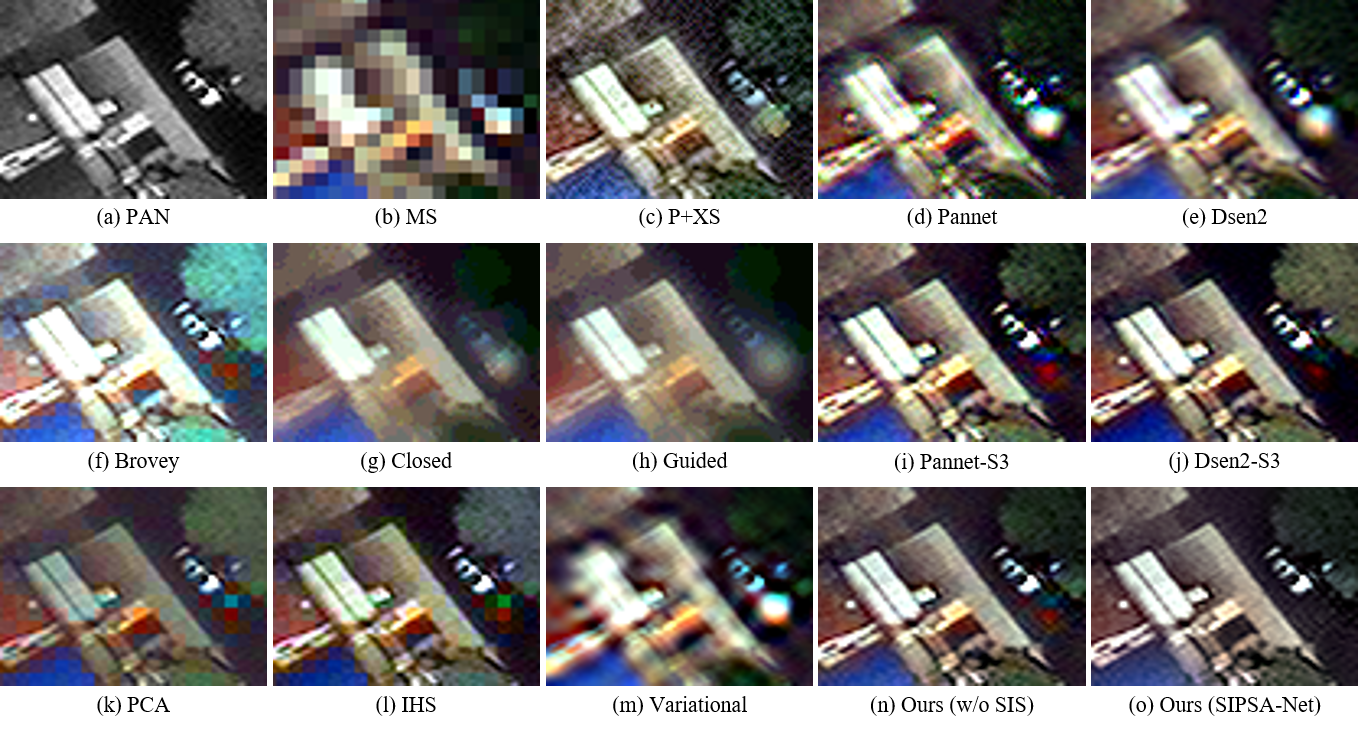}
  \caption{The figures are the pan-sharpened images by previous methods, our SIPSA-Net and SIPSA-Net without SiS loss function.}
  \label{fig:fig_6}
\end{figure*}
\paragraph{Quantitative comparisons}
Due to the unavailability of ground-truth pan-sharpened images, we evaluate the performances of SIPSA-Net and other PS methods under two different settings: lower-scale (input MS-scale) and full-scale (input PAN-scale) validations. For the lower-scale comparisons, the bilinear down-scaled PAN and MS images are used as input images. The resulting lower-scale output PS images are compared with their corresponding original-scale MS images in terms of three metrics: (i) spatial correlation coefficient (SCC) \cite{scc}; (ii) erreur relative global adimensionnelle de synthèse (ERGAS) \cite{ergas}; and (iii) peak signal-to-noise ratio (PSNR). SCC and ERGAS measure the spatial and spectral distortions, while PSNR measures PS images' reconstruction quality. However, the lower-scale comparison has a serious weak point that the lower-scale PS output cannot contain the original PAN details. 

On the other hand, under the full-scale validation, SCC$_{F}$ is measured between the original PAN inputs and the luminance channels of their PS output images \cite{scc}. SCC$_{F}$ indicates how much a pan-sharpening method can maintain the high-frequency components of the PAN input images in the PS output images. The quality-with-no-reference (QNR) \cite{qnr} metric, a no-reference metric for pan-sharpening, is also measured. The third metric of full scale evaluation is joint quality measure (JQM) \cite{jqm}. The JQM measures the degree of spectral distortion between MS and downscaled output images, and the spatial correlation coefficient between PAN and the output images. 
 
The proposed SIPSA-Net is designed to correct the inherent misalignment between the PAN and MS input images by aligning the color of objects in the MS input images with the corresponding details in the PAN input images. It is very important to note that for the PS output images aligned to the PAN input images, measuring the spectral distortion on the PS output images against the original MS input images is not appropriate since the original MS input images are not aligned with their corresponding PAN input images. Therefore, in addition to the conventional direct spectral distortion measures with original MS images, we also measure them with the aligned MS images produced by the correlation matching procedure (See the \textit{Supplemental} material associated with this paper for more details).

Table \ref{tab:metricQualitative comparisons_comparison} compares the performances of our SIPSA-Net and other SOTA methods in terms of SCC, ERGAS and PSNR in lower scale validations, and SCC$_F$, QNR and JQM in full-scale validations. Our SIPSA-Net trained with all features is denoted as SIPSA-Net (full), and the SIPSA-Net trained without the SiS loss function is denoted as SIPSA-Net (w/o SiS). As can be shown in Table \ref{tab:metricQualitative comparisons_comparison}, SIPSA-Net (full) and SIPSA-Net (w/o SiS) outperform all the previous methods in terms of SCC, ERGAS and PSNR, SCC$_F$ and JQM. However,the SIPSA-Net (full) shows relative lower performance than P+XS, PanNet, DSen2 \cite{pxs,pannet,dsen2} and SIPSA-Net (w/o SiS) in terms of QNR. It should be noted that QNR does not agree well with the human-perceived visual quality as discussed in \cite{multi, jqm, critical}. Instead, JQM \cite{jqm} was developed to improve QNR, which shows better agreement with human-perceived visual quality. Please see the detailed analysis on QNR and JQM in the \textit{Supplemental} material.

\paragraph{Qualitative comparisons}
Figs. \ref{fig:fig_4}, \ref{fig:fig_5} and \ref{fig:fig_6} show visual comparisons of PS results from our SIPSA-Net and the previous state-of-the-art methods. All three figures contain moving cars on the roads which cause a relatively large amount of misalignment between PAN and MS images. It can be seen in Figs. \ref{fig:fig_4}, \ref{fig:fig_5} and \ref{fig:fig_6} that the output PS images of all methods except for ours contain severe amount of artifacts around the cars. This shows the effectiveness of registration (alignment) learning by our FAM. It is also clearly shown in Fig. \ref{fig:fig_4}-(j) that our SIPSA-Net well preserves the high-frequency details of the PAN input and the spectral information of the MS input image in the PS image. In Fig. \ref{fig:fig_5}-(j), it can be seen that our SIPSA-Net can also produce the PS output with the highest visual quality compared to the other methods that suffer from severe color artifacts. Fig. \ref{fig:fig_6}-(n), and -(o) show the PS images of our SIPSA-Net trained without and with the SiS loss function, respectively. It is worthwhile to note that the PS image in Fig. \ref{fig:fig_6}-(n) has higher QNR score (0.867) than the PS image in Fig. \ref{fig:fig_6}-(o) (0.769) although it suffers from the color artifact (a big red spot as shown in Fig. \ref{fig:fig_6}-(n)). However, the JQM value of Fig. \ref{fig:fig_6}-(o) (0.928) is higher than that of Fig. \ref{fig:fig_6}-(n) (0.904) , which is consistent with the perceived visual quality. 
\subsection{Ablation study}
\label{sec:ablation}
Ablation studies have been conducted in three different conditions to show the effectiveness of key aspects of our proposed SIPSA-Net. Throughout the experiments, the SIPSA-Net has been inspected by removing each key component from its full version. The resulting different versions (models) have been evaluated in the full-scale using the original MS and PAN images as inputs for the networks. We use three metrics to measure the performance of the different versions: SCC$_F$\cite{scc}, ERGAS\cite{ergas}, and JQM \cite{jqm}. 

Table \ref{tab:Ablation} provides the performance of the different versions with/without the SiS loss and FAM. Also, there is a version that applies PWPAC directly in pixel domain for MS input images, called 'SIPSA-Net w/ PAM (pixel alignment module)'. The 'SIPSA-Net w/o SiS loss' version is trained without the SiS loss, and the FAM is excluded for the 'SIPSA-Net w/o FAM' version that directly takes the original MS images as input without alignment. As can be seen in Table \ref{tab:Ablation}, 
the full version of the SIPSA-Net shows the best performance in terms of JQM. The 'SIPSA-Net w/o SiS loss' version is superior to all the other versions in terms of SCC$_F$ and ERGAS and the full version is the second best. Without the SiS loss, the SISPA-Net suffers from various artifacts such as comet tails, false color and misalignment as in Fig. \ref{fig:fig_6}-(n) where the big red and blue spots (color artifacts) nearby the car appear.
It is also worthwhile to mention that for the 'SISPA-Net w/o FAM' version that the FAM plays a crucial role in generating the PS images of high visual quality as shown in Table \ref{tab:Ablation} (see also Fig. 3 in the \textit{Supplemental} material).
Lastly, the 'SIPSA-Net w/ PAM' version is difficult to be well trained because the MS images and their corresponding PAN images have different signal characteristics and are difficult to be directly aligned in pixel domain rather than in feature domain.

\begin{table}[!]
    \small
    \centering
    \setlength{\tabcolsep}{5pt}
    \caption{Ablation study on three different conditions.}
    \input{Ablation_study}
    \label{tab:Ablation}

\end{table}

\section{Conclusions}
Our SIPSA-Net is the first work that can effectively handle both global and local misalignment in MS and PAN image pairs. For this, we propose a novel feature alignment module (FAM) to obtain aligned MS images. The FAM yields aligned MS images to the PAN images by performing pixel-wise probabilistic alignment convolutions (PWPAC) for the extracted MS features with an estimated pixel-wise offset probability map (PWOPM). Then, the following pan-sharpening module (PSM) can effectively generate the PS images of high quality. To effectively train SIPSA-Net robust to the misaligned MS and PAN images, we propose two loss functions: an edge detail loss between PS outputs and PAN images and a  shift-invariant-spectral (SiS) loss between the PS outputs and the input MS images. The edge detail loss helps the SIPSA-Net effectively capture the details of PAN images into PS output generation while the SiS loss allows the SIPSA-Net to learn the alignment between misaligned MS and PAN images in generating the PS output stably aligned to the PAN images even for the fast moving objects. Extensive experiments show that our SIPSA-Net significantly outperforms the SOTA methods quantitatively and qualitatively. 

\small{
\smallskip\noindent
\textbf{Acknowledgement.}\quad
This research was supported by the MSIT (Ministry of Science and ICT), Korea, under the ITRC (Information Technology Research Center) support program (IITP-2021-2016-0-00288) supervised by the IITP (Institute for Information \& Communications Technology Planning \& Evaluation).}

{\small
\bibliographystyle{ieee_fullname}

\input{main.bbl}
}

\end{document}

%% file: with_psnr.tex
\def\arraystretch{1.2}
\begin{tabular}{l|ccc|ccc|ccc}
\hline
 & \multicolumn{3}{c|}{Lower-scale (with aligned MS)} & \multicolumn{3}{c|}{Full-scale (with aligned MS)} & \multicolumn{3}{c}{Lower-scale (with misaligned MS)}\\ \cline{2-10}

 & $SCC$ $\uparrow$ & $ERGAS$ $\downarrow$ & $PSNR$ $\uparrow$ & $SCC_{F}$ $\uparrow$ & $QNR$ $\uparrow$ & $JQM$ $\uparrow$ & $SCC$ $\uparrow$ & $ERGAS$ $\downarrow$ & $PSNR$ $\uparrow$\\
\hline
Brovey \cite{brovey} & 0.932 &6.129&27.655 & 0.952 & 0.780 & 0.884 & 0.836 & 7.190  & 26.098 \\ 
Affinity \cite{affinity} & 0.873 & 3.453 & 33.264 & 0.747 & 0.677 & 0.759  & 0.791 & 3.993  & 32.207\\
GF \cite{guided} & 0.886 & 3.394 & 33.376 & 0.778 & 0.680 & 0.761  & 0.801 & 4.005 &  32.175\\
IHS \cite{ihs} & 0.928 & 2.763 & 35.391 & 0.958 & 0.852 & 0.883 & 0.836 & 3.877 & 32.676  \\
PCA \cite{pca} & 0.908 & 4.575 & 31.187  & 0.899 & 0.743 & 0.845 & 0.817 & 5.060  & 30.41\\
P+XS \cite{pxs}& 0.898 & 3.402 & 32.553 & 0.858 & 0.851 & 0.863 & 0.775 & 4.459 & 30.365\\
Variational \cite{variational} & 0.839 & 2.931 & 35.379 & 0.665 & 0.754 & 0.752 & 0.781 & \underline{3.638} & \underline{33.415} \\
\hline
PNN \cite{pnn} & 0.762 & 5.111  & 30.612 & 0.697 & 0.649 & 0.752 & 0.694 & 5.694 & 29.594\\
PanNet \cite{pannet} & 0.877 & 3.315 & 34.113 & 0.811 & 0.806 & \underline{0.916} & 0.802 & 4.047 & 32.374\\
PanNet-S3 \cite{s3} & 0.933 & 2.724 & 35.454 & 0.954 & 0.844 & 0.827 & 0.838 & 3.816 & 32.769 \\
DSen2 \cite{dsen2} & 0.820  & 3.763 & 32.939 & 0.837 & \underline{0.852} & 0.908 & 0.820  & 3.763 & 32.939 \\
DSen2-S3 \cite{s3} & 0.829 & 2.931 & 35.057  & {0.959} & 0.808 & 0.758 & \textbf{0.844} & 5.330  & 29.39\\
\hline
SIPSA-Net (full) & \textbf{0.940} & \underline{2.500} & \underline{36.354}  & \underline{0.962}&  0.798 & \textbf{0.934} & 0.842 & 3.683 & 33.216\\	
SIPSA-Net (w/o SiS) & \textbf{0.940} & \textbf{2.351} & \textbf{36.905} & \textbf{0.964}&  \textbf{0.858} & 0.906 & \textbf{0.844} & \textbf{3.505} & \textbf{33.662}\\	

\hline
\end{tabular}

%% file: Ablation_study.tex
\def\arraystretch{1.2}
\begin{tabular}{l|ccccc|ccc}
\hline

 & $SCC_F$ $\uparrow$ & $ERGAS$ $\downarrow$ & $JQM$ $\uparrow$ \\
\hline
SIPSA-Net (full)  & \underline{0.962} & \underline{2.500} & \textbf{0.934} \\

\hline
SIPSA-Net w/o SiS loss  & \textbf{0.964} & \textbf{2.351} & \underline{0.906}  \\
\hline
SIPSA-Net w/o FAM  & 0.340 & 4.067 & 0.576  \\
\hline
SIPSA-Net w/ PAM  & 0.318 & 4.094 & 0.575  \\
\hline
\end{tabular}

%% file: main.bbl
\begin{thebibliography}{10}\itemsep=-1pt

\bibitem{tensorflow}
Mart{\'\i}n Abadi, Paul Barham, Jianmin Chen, Zhifeng Chen, Andy Davis, Jeffrey
  Dean, Matthieu Devin, Sanjay Ghemawat, Geoffrey Irving, Michael Isard, et~al.
\newblock Tensorflow: A system for large-scale machine learning.
\newblock In {\em 12th $\{$USENIX$\}$ Symposium on Operating Systems Design and
  Implementation ($\{$OSDI$\}$ 16)}, pages 265--283, 2016.

\bibitem{qnr}
Luciano Alparone, Bruno Aiazzi, Stefano Baronti, Andrea Garzelli, Filippo
  Nencini, and Massimo Selva.
\newblock Multispectral and panchromatic data fusion assessment without
  reference.
\newblock {\em Photogrammetric Engineering \& Remote Sensing}, 74(2):193--200,
  2008.

\bibitem{Regularize}
H.~A. {Aly} and G. {Sharma}.
\newblock A regularized model-based optimization framework for pan-sharpening.
\newblock {\em IEEE Transactions on Image Processing}, 23(6):2596--2608, 2014.

\bibitem{pxs}
Coloma Ballester, Vicent Caselles, Laura Igual, Joan Verdera, and Bernard
  Roug{\'e}.
\newblock A variational model for p+ xs image fusion.
\newblock {\em International Journal of Computer Vision}, 69(1):43--58, 2006.

\bibitem{ihs}
WJOSEPH CARPER, THOMASM LILLESAND, and RALPHW KIEFER.
\newblock The use of intensity-hue-saturation transformations for merging spot
  panchromatic and multispectral image data.
\newblock {\em Photogrammetric Engineering and remote sensing}, 56(4):459--467,
  1990.

\bibitem{s3}
Jae-Seok Choi, Yongwoo Kim, and Munchurl Kim.
\newblock S3: A spectral-spatial structure loss for pan-sharpening networks.
\newblock {\em IEEE Geoscience and Remote Sensing Letters}, 2019.

\bibitem{srcnn}
Chao Dong, Chen~Change Loy, Kaiming He, and Xiaoou Tang.
\newblock Image super-resolution using deep convolutional networks.
\newblock {\em IEEE transactions on pattern analysis and machine intelligence},
  38(2):295--307, 2015.

\bibitem{variational}
Xueyang Fu, Zihuang Lin, Yue Huang, and Xinghao Ding.
\newblock A variational pan-sharpening with local gradient constraints.
\newblock In {\em Proceedings of the IEEE Conference on Computer Vision and
  Pattern Recognition}, pages 10265--10274, 2019.

\bibitem{brovey}
Alan~R Gillespie, Anne~B Kahle, and Richard~E Walker.
\newblock Color enhancement of highly correlated images. i. decorrelation and
  hsi contrast stretches.
\newblock {\em Remote Sensing of Environment}, 20(3):209--235, 1986.

\bibitem{multi}
Manjunath~V Joshi and Kishor~P Upla.
\newblock {\em Multi-resolution Image Fusion in Remote Sensing}.
\newblock Cambridge University Press, 2019.

\bibitem{vdsr}
Jiwon Kim, Jung Kwon~Lee, and Kyoung Mu~Lee.
\newblock Accurate image super-resolution using very deep convolutional
  networks.
\newblock In {\em Proceedings of the IEEE conference on computer vision and
  pattern recognition}, pages 1646--1654, 2016.

\bibitem{dsen2}
Charis Lanaras, Jos{\'e} Bioucas-Dias, Silvano Galliani, Emmanuel Baltsavias,
  and Konrad Schindler.
\newblock Super-resolution of sentinel-2 images: Learning a globally applicable
  deep neural network.
\newblock {\em ISPRS Journal of Photogrammetry and Remote Sensing},
  146:305--319, 2018.

\bibitem{edsr}
Bee Lim, Sanghyun Son, Heewon Kim, Seungjun Nah, and Kyoung Mu~Lee.
\newblock Enhanced deep residual networks for single image super-resolution.
\newblock In {\em Proceedings of the IEEE conference on computer vision and
  pattern recognition workshops}, pages 136--144, 2017.

\bibitem{adamw}
Ilya Loshchilov and Frank Hutter.
\newblock Decoupled weight decay regularization.
\newblock {\em arXiv preprint arXiv:1711.05101}, 2017.

\bibitem{decomposition}
Stephane~G Mallat.
\newblock A theory for multiresolution signal decomposition: the wavelet
  representation.
\newblock {\em IEEE transactions on pattern analysis and machine intelligence},
  11(7):674--693, 1989.

\bibitem{pnn}
Giuseppe Masi, Davide Cozzolino, Luisa Verdoliva, and Giuseppe Scarpa.
\newblock Pansharpening by convolutional neural networks.
\newblock {\em Remote Sensing}, 8(7):594, 2016.

\bibitem{jqm}
Gintautas Palubinskas.
\newblock Joint quality measure for evaluation of pansharpening accuracy.
\newblock {\em Remote Sensing}, 7(7):9292--9310, 2015.

\bibitem{pca}
Vijay~P Shah, Nicolas~H Younan, and Roger~L King.
\newblock An efficient pan-sharpening method via a combined adaptive pca
  approach and contourlets.
\newblock {\em IEEE transactions on geoscience and remote sensing},
  46(5):1323--1335, 2008.

\bibitem{undecimated}
Jean-Luc Starck, Jalal Fadili, and Fionn Murtagh.
\newblock The undecimated wavelet decomposition and its reconstruction.
\newblock {\em IEEE transactions on image processing}, 16(2):297--309, 2007.

\bibitem{pac}
H. {Su}, V. {Jampani}, D. {Sun}, O. {Gallo}, E. {Learned-Miller}, and J.
  {Kautz}.
\newblock Pixel-adaptive convolutional neural networks.
\newblock In {\em 2019 IEEE/CVF Conference on Computer Vision and Pattern
  Recognition (CVPR)}, pages 11158--11167, 2019.

\bibitem{affinity}
Stephen Tierney, Junbin Gao, and Yi Guo.
\newblock Affinity pansharpening and image fusion.
\newblock In {\em 2014 International Conference on Digital Image Computing:
  Techniques and Applications (DICTA)}, pages 1--8. IEEE, 2014.

\bibitem{guided}
Kishor~P Upla, Sharad Joshi, Manjunath~V Joshi, and Prakash~P Gajjar.
\newblock Multiresolution image fusion using edge-preserving filters.
\newblock {\em Journal of Applied Remote Sensing}, 9(1):096025, 2015.

\bibitem{critical}
Gemine Vivone, Luciano Alparone, Jocelyn Chanussot, Mauro Dalla~Mura, Andrea
  Garzelli, Giorgio~A Licciardi, Rocco Restaino, and Lucien Wald.
\newblock A critical comparison among pansharpening algorithms.
\newblock {\em IEEE Transactions on Geoscience and Remote Sensing},
  53(5):2565--2586, 2014.

\bibitem{ergas}
Lucien Wald.
\newblock {\em Data fusion: definitions and architectures: fusion of images of
  different spatial resolutions}.
\newblock Presses des MINES, 2002.

\bibitem{pannet}
Junfeng Yang, Xueyang Fu, Yuwen Hu, Yue Huang, Xinghao Ding, and John Paisley.
\newblock Pannet: A deep network architecture for pan-sharpening.
\newblock In {\em Proceedings of the IEEE International Conference on Computer
  Vision}, pages 5449--5457, 2017.

\bibitem{rcan}
Yulun Zhang, Kunpeng Li, Kai Li, Lichen Wang, Bineng Zhong, and Yun Fu.
\newblock Image super-resolution using very deep residual channel attention
  networks.
\newblock In {\em Proceedings of the European Conference on Computer Vision
  (ECCV)}, pages 286--301, 2018.

\bibitem{bdpn}
Yongjun Zhang, Chi Liu, Mingwei Sun, and Yangjun Ou.
\newblock Pan-sharpening using an efficient bidirectional pyramid network.
\newblock {\em IEEE Transactions on Geoscience and Remote Sensing},
  57(8):5549--5563, 2019.

\bibitem{scc}
J Zhou, DL Civco, and JA Silander.
\newblock A wavelet transform method to merge landsat tm and spot panchromatic
  data.
\newblock {\em International journal of remote sensing}, 19(4):743--757, 1998.

\end{thebibliography}
